# An Information Theory For Preferences

Ali E. Abbas

*Department of Management Science and Engineering, Stanford University, Stanford, Ca, 94305*

**Abstract.** Recent literature in the last Maximum Entropy workshop introduced an analogy between cumulative probability distributions and normalized utility functions [1]. Based on this analogy, a utility density function can de defined as the derivative of a normalized utility function. A utility density function is non-negative and integrates to unity. These two properties of a utility density function form the basis of a correspondence between utility and probability, which allows the application of many tools from one domain to the other. For example, La Place's principle of insufficient reason translates to a principle of insufficient preference. The notion of uninformative priors translates to uninformative utility functions about a decision maker's preferences. A natural application of this analogy is a maximum entropy principle to assign maximum entropy utility values. Maximum entropy utility interprets many of the common utility functions based on the preference information needed for their assignment, and helps assign utility values based on partial preference information. This paper reviews maximum entropy utility, provides axiomatic justification for its use, and introduces further results that stem from the duality between probability and utility, such as joint utility density functions, utility inference, and the notion of mutual preference.

## INTRODUCTION

In many decision situations, we are faced with multiple and conflicting attributes. When the decision situation is deterministic, each decision alternative is described by a single prospect (consequence). The problem of choosing the best decision alternative is that of ordering the prospects present or, alternatively, assigning a value function over the attributes of each prospect. The optimal decision alternative is the one that has the largest value as determined by the value function or the highest order in the ranked list. A normative justification for the order requirement is that a decision maker who cannot order the prospects is vulnerable to being a "money pump" (we would choose prospect A over prospect B, but also prospect B over prospect A, and be willing to pay money to move from one prospect to the other)

When uncertainty is present, a decision alternative is characterized by several prospects and a probability distribution representing the probability of their occurrence. In this case, the rank order of the prospects alone is insufficient to determine the optimal decision alternative, and the Von Neumann and Morgenstern utility values need to be elicited. [2]. To elicit these utility values, a decision maker is first asked to order the prospects of the decision situation from best to worst. Once the order is provided, the next step is to assign a utility value for each prospect. For any three ordered prospects, $P_1 \succ P_2 \succ P_3$, the decision maker assigns a probability, $\pi$, for which she is indifferent to receiving $P_2$ for sure and a deal where she would receive $P_1$ with probability $\pi$ and $P_3$ with probability $(1-\pi)$. If the utility values of $P_1$ and $P_3$ are one and zero respectively, then the utility value of $P_2$, also called the preference probability of $P_2$, will be equal to $\pi(1)+(1-\pi)(0)=\pi$. The higher the prospect is in the ranked list, the larger is the utility value assigned to it. The optimal decision alternative is now the one, which has the highest expected utility for its prospects.

The normalization of the utility values discussed above leads to an analogy between utility functions and cumulative probability distributions. Our goal in this paper is to review the analogy between utility and probability, and to draw on the applications of this analogy to the maximum entropy principle and to many other concepts of information theory. The main requirement for this analogy is that the decision maker can provide the complete preference order for the prospects of the decision situation she is facing.

## UTILITY ASSIGNMENT FOR ORDERED PROSPECTS

We start this section with the definition of a utility vector for a set of *K* ordered prospects of a decision situation. A utility vector contains the utility values of the prospects starting from lowest to highest. With no significant loss of generality, we will assign a utility value of zero to the least preferred prospect, $u_0$, and a value of one to the most preferred prospect, $u_{K-1}$ (we ignore the case of absolute indifference between the *K* prospects). The utility vector has K elements defined as

$$U \triangleq (u_0, u_1, u_2, \ldots, u_{K-2}, u_{K-1}) = (0, u_1, u_2, \ldots, u_{K-2}, 1). \qquad (1)$$

Any utility vector of dimension K can be represented as a point in a K-2 dimensional space in the region defined by $0 \leq u_1 \leq u_2 \leq \ldots \leq u_{K-3} \leq u_{K-2} \leq 1$. We will call this region the utility volume. An example of a two-dimensional utility volume is shown in figure (1).

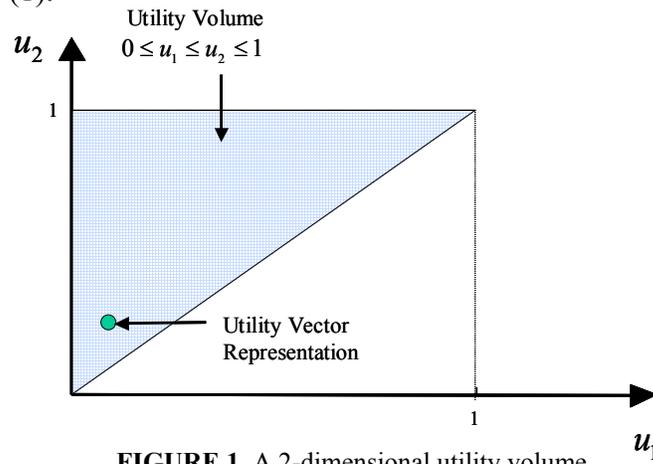

**FIGURE 1.** A 2-dimensional utility volume.

The second definition is the utility-increment vector, $\Delta U$, whose elements are equal to the difference between the consecutive elements in the utility vector. The utility-increment vector has $(K-1)$ elements defined as

$$\Delta U \triangleq (u_1 - 0, u_2 - u_1, ......, 1 - u_{K-2}) = (\Delta u_1, \Delta u_2, \Delta u_3, ...., \Delta u_{K-1}). \quad (2)$$

Note that all elements of $\Delta U$ are greater than or equal to zero and sum to one. Therefore, any utility-increment vector can be represented as a point in a (K-1) dimensional simplex $\left\{ x : \sum_{i=1}^{K-1} x_i = 1, x_i \geq 0 \right\}$. These two properties of the utility increment vector over the simplex form the basis of the analogy between probability and utility and will be used further in this paper. We will refer to this simplex as the utility simplex. For example, a 3-dimensional utility simplex is shown in figure (2).

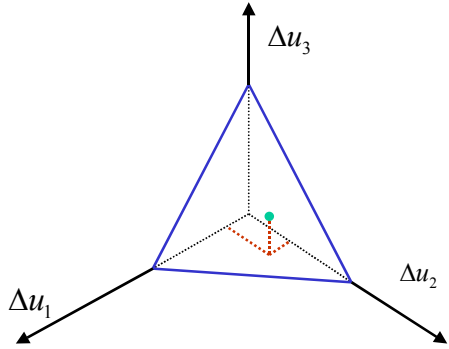

**FIGURE 2.** A 3-dimensional utility simplex.

Note that all points in the utility volume or the utility simplex satisfy the decision maker's preference ordering of the prospects but assign different utility values to them. In other words, knowledge of the preference order alone tells us nothing about the location of the utility vector over the utility volume or the utility increment vector over the utility simplex. If all we know about the decision maker's preferences is the ordering of the prospects, it is reasonable to assume that the location of the utility increment vector is uniformly distributed over the utility simplex. This is also the maximum entropy distribution over the simplex. The uniform distribution over the utility simplex implies that the location of the utility increment vector is determined by a Dirichlet distribution whose K-1 parameters are equal to one. This is in turn implies the following three propositions described in [1].

## Proposition 1: Marginal Distributions of the Utility Vector

The maximum entropy marginal distributions for the utility values of a set of K ordered prospects are the family of Beta distributions, $Beta(j, K - j - 1), \ j = 1,....K - 2$.

Given the marginal distribution for each element in the utility vector, it is known that the logical assignment of each utility increment value is the mean of its marginal distribution [3]. The mean of a $Beta(j, K-j-1)$ distribution is equal to $\frac{j}{K-1}$, and is the utility value you would assign to each prospect, j. The utility values for prospects j=0 and j=K-1 are deterministic with values 0 and 1 respectively.

## Proposition 2: Marginal Distributions of the Utility Increment Vector

The maximum entropy marginal distributions for the elements of the utility increment vector have identical maximum entropy marginal distributions, $Beta(1, K-2)$.

Given the marginal distribution for each element in the utility increment vector, it is known that the logical assignment of each utility increment value is the mean of its marginal distribution [3]. The mean of a $Beta(1, K-2)$ distribution is $\frac{1}{K-1}$. We now have a maximum entropy assignment for the location of the utility increment vector and are ready for proposition 3 below.

## Proposition 3: Utility Increment Vector Assignment

The maximum entropy assignment for increments in the utility values of a set of $K$ ordered prospects yields identical increments in utility values equal to $\frac{1}{K-1}$.

For example, if we have a decision situation with five ordered prospects, $A \succ B \succ C \succ D \succ E$, the maximum entropy assignment of the utility increment vector is $\Delta U = (0.25, 0.25, 0.25, 0.25)$ and the maximum entropy assignment of the utility vector is $U = (0, 0.25, 0.5, 0.75, 1)$. Note that using the definitions of utility volume and utility simplex, as well as the maximum entropy uniform joint distribution, we have made an unbiased assignment of utility values for five prospects when only the preference order is known. Note also that this unbiased assignment gave equal increments in utility values.

Now we are ready for the first application of our utility-probability analogy: to assign unbiased utility values when only the order of the prospects is available. We mean by "unbiased" utility values, those that do not lead to arbitrary assumptions of preference information that is not available.

## THE PRINCIPLE OF INSUFFICIENT PREFERENCE

In the previous section we illustrated how to assign utility values for K prospects of a decision situation when only the preference order is available. Now we show how to work through the same problem using our utility-probability analogy. The analogous problem in probability land is to assign a probability to the outcomes of a discrete random variable when no further information is available. This problem dates back to Laplace's principle of insufficient reason. Laplace suggested that we assign equal probabilities to all outcomes unless there is information that suggests otherwise. The utility-probability analogy then suggests a principle of insufficient preference, which can be stated as follows: when only the preference order of the prospects is known, we assign equal increments in utility values unless there is preference information that suggests otherwise. Note that the principle of insufficient preference provides the same assignment of utility values as suggested by proposition 3 above.

The principle of insufficient preference assigns utility values for discrete ordered prospects. In many cases, however, we would like to assign utility values over a continuous domain. In the next section we will extend our analysis to the continuous case and present an additional element to our utility probability analogy: a utility density function.

## UTILITY FUNCTIONS AND UTILITY DENSITY FUNCTIONS

In the continuous case the utility vector has an infinite number of elements tracing a utility curve over the domain of monetary prospects. The largest value is still one and the lowest value is zero corresponding to the utility values of the best and least preferred prospects respectively. A utility curve thus has the same mathematical properties as a cumulative distribution function. The utility-increment vector is now the derivative of the utility curve and represents a "utility density function", $u(x),$ which is analogous to a probability density function in that it is non-negative and integrates to unity.

$$u(x) \triangleq \frac{\partial U(x)}{\partial x} \qquad (3)$$

Our problem of assigning utility values for continuous prospects of a decision situation, when partial preference information is available, is now equivalent to that of assigning an unbiased utility density function for the ordered prospects. The latter problem is analogous to the problem of assigning prior probabilities to a continuous random variable when only partial information is available.

In this paper we will generalize the utility assignment problem and present a maximum entropy approach for the assignment of utility values when only partial preference information is available. First let us discuss the entropy of a utility function and its interpretation.

# THE ENTROPY OF A UTILITY FUNCTION

Having defined the notion of a utility density function, it is natural to extend our analysis to the entropy measure applied to a utility function, and discuss its interpretations. The differential entropy term applied to a utility density function is

$$h(x) = -\int u(x)\ln(u(x))dx \qquad (4)$$

Let us take, as an example, an exponential utility density function over a bounded domain [a, b]. We have

$$u(x) = \frac{\gamma e^{-\gamma x}}{e^{-\gamma a} - e^{-\gamma b}}, \ a \leq x \leq b \qquad (5)$$

where $\gamma$ is the decision maker's risk aversion coefficient.

As $\gamma \to 0$, the decision maker becomes risk neutral (he values uncertain deals at their expected value). Using L'hopital's formula, we can also show that

$$\frac{\gamma e^{-\gamma x}}{e^{-\gamma a} - e^{-\gamma b}} \to \frac{1}{b-a} \text{ as } \gamma \to 0. \qquad (6)$$

The utility density function approaches a uniform density and the entropy of the utility function thus achieves its maximum value of $\ln(b-a)$. Note that the uniform utility density function matches the principle of insufficient preference discussed earlier as it provides equal increments in utility values for equally spaced prospects. The utility function that corresponds to this uniform utility density function can be obtained by integration and is linear.

The linear utility function is thus the maximum entropy utility function over a continuous bounded domain.

As $\gamma \to +\infty$,

$$\frac{\gamma e^{-\gamma x}}{e^{-\gamma a} - e^{-\gamma b}} \to \delta(x-a) \qquad (7)$$

The utility density function converges to an impulse utility density function and the entropy approaches a minimum value of negative infinity showing less "spread" in the utility density. The impulse utility density function is the minimum entropy utility function and integrates to a step utility function.

As $\gamma \to -\infty$,

$$\frac{\gamma e^{-\gamma x}}{e^{-\gamma a} - e^{-\gamma b}} \to \delta(x-b) \qquad (8)$$

The utility density function converges again to an impulse utility density function, and the entropy approaches a minimum value of negative infinity. This is the other extreme where the step in the utility function occurs at the upper bound.

The step utility function is thus the minimum entropy utility function and corresponds, in this example, to cases of extreme risk aversion or risk seeking behavior.

Figure 3 shows an example of two exponential utility density functions for different values of $\gamma$ and different entropy values.

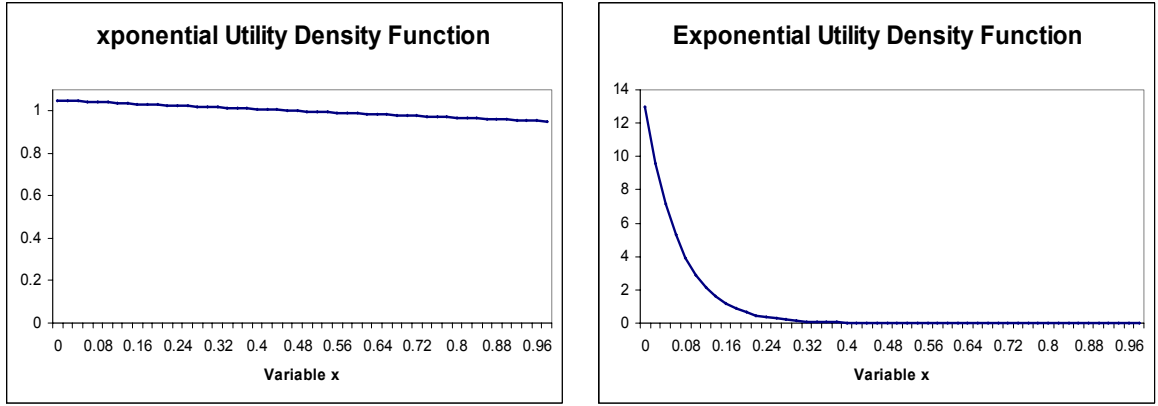

**FIGURE 3**. (a) Utility density function for small values of $\gamma$ is approximately uniform showing large spread. (b) Utility density function for large values of $\gamma$ converges to an impulse utility density showing less spread and integrates to a step utility function.

A step utility function appears often in behavioral decision making literature and is known as the aspiration utility function. Simon [4] suggested that individuals could simplify decision problems by having binary goals: satisfactory if the outcome is above the aspiration level or unsatisfactory if it is below it.

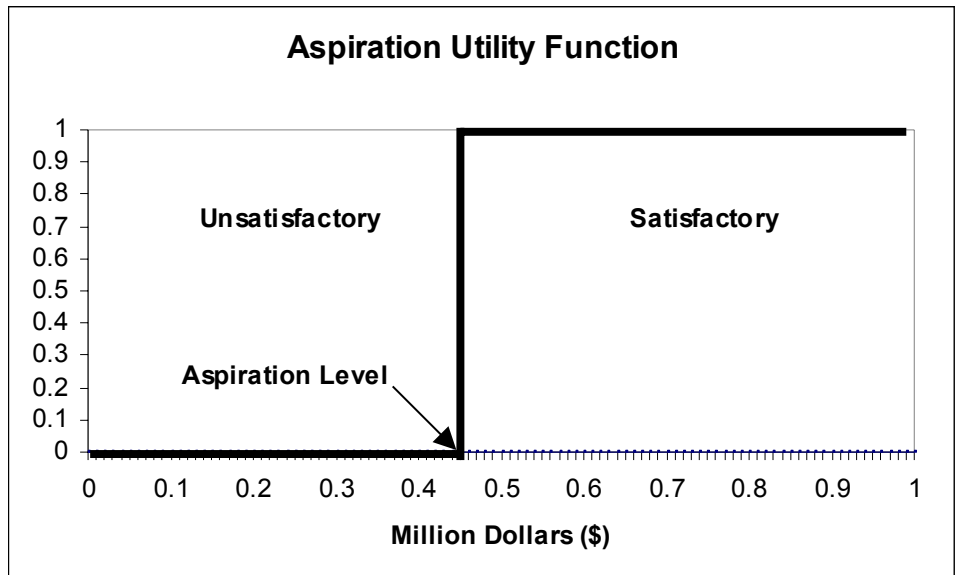

**FIGURE 4.** Aspiration (Minimum Entropy) Utility Function

The interpretation of entropy for utility functions is thus a measure of spread about a decision maker's aspiration level. Aspiration utility functions have the least entropy in their utility density functions showing binary preferences, while uniform utility density functions have the highest entropy showing smoother utility functions that vary gradually over a bounded domain.

## MAXIMUM ENTROPY UTILITY FUNCTIONS

Now we will make use of the concept of the entropy of a utility function and present a method to assign "unbiased" utility values based on the partial preference information we know about the decision maker. By partial preference information, we mean any information that includes the complete preference order of the prospects but does not include knowledge of the complete utility values.

In the probability domain, Edwin Jaynes [5] proposed the maximum entropy principle suggesting the use of a probability distribution, which has maximum entropy subject to whatever information is known. In an analogous manner, we defined the maximum entropy utility principle, [1], that suggests the use of the utility function (or utility vector) whose utility density function (or utility increment vector) has maximum entropy subject to whatever preferences are known. This result is an extension of the principle of insufficient preference described earlier.

Maximum entropy utility provides an interpretation for many of the utility functions used in practice based on the preference information constraints needed for their assignment. For example when only bounds on the domain of continuous outcomes are available, the maximum entropy utility density function is uniform and integrates to a risk neutral utility function as we have discussed earlier. When the first moment of the utility function is known over a positive domain, the maximum entropy utility function is exponential with the first moment equal to the risk tolerance. This provides us with a new interpretation for the first moment of an exponential utility function over a positive domain. When the first and second moments are known, the maximum entropy utility function over a continuous domain leads to the famous S-shaped prospect theory utility function [6].

Maximum entropy utility presents a method to assign unbiased utility values when partial preference information is available. For example, the maximum entropy utility function given knowledge of some utility values leads to a piecewise linear utility function that passes through the given utility points.

## UNINFORMATIVE UTILITY FUNCTIONS

Much of the research on uninformative and entropic priors for probability functions can be translated through our utility-probability analogy into similar research on uninformative utility functions. For example, when Jeffreys' [7] uninformative prior, $u(x) = \frac{1}{x}, x > 0$ is used as an uninformative utility density function, it leads to the famous logarithmic utility function that is used very often to represent the decision maker's preferences in the literature.

## ATTRIBUTE DOMINANCE UTILITY FUNCTIONS

In this section we present the extension of the previous analysis to multiattribute utility functions with ordered attributes. We assume that $(x_{\min}, y_{\min})$ is the least preferred prospect and $(x_{\max}, y_{\max})$ is most preferred. Furthermore we will assume that the values of $(x, y)$ are arranged such that for any $x_0, x_1, y_0, y_1$ we have

$$x_1 > x_0 \Rightarrow (x_1, y) \succ (x_0, y) \forall y \text{ and } y_1 > y_0 \Rightarrow (x, y_1) \succ (x, y_0) \forall x \qquad (9)$$

With no loss of generality, we will use a normalized multiattribute utility function, $U_{xy}(x, y)$, over the attributes in all of our analysis. i.e.

$$0 \leq U_{xy}(x, y) \leq 1 \qquad (10)$$

Based on this formulation, we have

$$U_{xy}(x_{\min}, y_{\min}) = 0, \ U_{xy}(x_{\max}, y_{\max}) = 1 \qquad (11)$$

Now we will focus on a class of multiattribute utility functions where a prospect $(x, y)$ is a least preferred prospect if either $x$ or $y$ is at its minimum value. This requirement places the following constraints on the utility function:

$$U_{xy}(x_{\min}, y_{\min}) = U_{xy}(x_{\min}, y) = U_{xy}(x, y_{\min}) = 0 \quad \forall x \in [x_{\min}, x_{\max}], y \in [y_{\min}, y_{\max}] \qquad (12)$$

We will call the multiattribute utility functions that satisfy condition (12) attribute dominance utility functions since any attribute set at a minimum dominates the remaining attributes and sets the multiattribute utility function to a minimum.

For attribute dominance utility functions we will define the marginal utility function over a single attribute, $x$, as the numerical value of the multiattribute utility function when all other attributes are set at their maximum values. I.e. for two attributes we have

$$U_x(x) \triangleq U_{xy}(x, y_{\max}) \qquad (13)$$

Now we define a conditional utility function, $U_{y|x}(y | x)$, as

$$U_{y|x}(y | x) \triangleq \frac{U_{xy}(x, y)}{U_x(x)} = \frac{U_{xy}(x, y)}{U_{xy}(x, y_{\max})} \qquad (14)$$

A conditional utility function, $U_{y|x}(y|x)$, is the utility function for attribute $y$ when we are guaranteed a fixed amount of attribute $x$. The conditional utility function $U_{y|x}(y|x)$ is the path traced on the multiattribute utility function for different values of $y$ and a fixed value of $x$. Equation (14) shows that for attribute dominance utility functions, the minimum value of the conditional utility function, $U_{y|x}(y_{\min}|x)$, is equal to zero and the maximum value, $U_{y|x}(y_{\max}|x)$, is equal to one. The denominator in equation (14) thus serves as a normalizing term so a conditional utility function can be assessed directly from the decision maker on a scale of 0 to 1.

Re-arranging equation (14) gives:

$$U_{xy}(x,y) = U_x(x)U_{y|x}(y|x) \tag{15}$$

In a similar manner, we can also define $U_{x|y}(x|y)$ as

$$U_{x|y}(x|y) \triangleq \frac{U_{xy}(x,y)}{U_y(y)} \tag{16}$$

Combining equations (14) and (16), we have

$$U_{x|y}(x|y) = \frac{U_{y|x}(y|x)U_x(x)}{U_y(y)} \tag{17}$$

Equation (17) provides a method to update our utility values over one attribute when we are guaranteed a fixed amount of another. We will call equation (17) Bayes' rule for utility inference.

Now define a joint utility density function, $u_{xy}(x,y)$, over two attributes, $x \in [x_{\min}, x_{\max}]$ and $y \in [y_{\min}, y_{\max}]$ as:

$$u_{xy}(x,y) \triangleq \frac{\partial^2 U(x,y)}{\partial x \partial y} \tag{18}$$

where $U_{xy}(x,y)$ is the multiattribute utility function for the two attributes $x$ and y.

From the properties of derivatives, the integral of a joint utility density function that is defined by equation (18) over a region $a \le x \le b, c \le y \le d$ is

$$\int_c^d \int_a^b u_{xy}(x,y)dxdy = U_{xy}(b,d) - U_{xy}(a,d) - U_{xy}(b,c) + U_{xy}(a,c) \tag{19}$$

If we set $a = x_{\min}$ and $c = y_{\min}$, and if attribute dominance utility exists, then equation (19) reduces to

$$U_{xy}(x,y) = \int_{y_{\min}}^{y} \int_{x_{\min}}^{x} u_{xy}(x,y)dxdy \qquad (20)$$

Equation (20) illustrates how attribute dominance utility functions can be obtained by integrating a joint utility density function from the least preferred prospect to a prospect $(x,y)$. This integral provides an analogy with joint cumulative distribution functions obtained by integrating a joint probability density function over the domain of the variables.

***We note that any general multiattribute utility function can be decomposed into a linear combination of attribute dominance utility functions.*** For example, rearranging equation (19) and using any general prospect $b=x$, $d=y$, $a=x_{\min}$ and $c=y_{\min}$ gives

$$U_{xy}(x,y) = U_{xy}(x,y_{\min}) + U_{xy}(x_{\min},y) - U_{xy}(x_{\min},y_{\min}) + \int_{y_{\min}}^{y} \int_{x_{\min}}^{x} u_{xy}(x,y)dxdy \qquad (21)$$

The terms $U_{xy}(x,y_{\min})$ and $U_{xy}(x_{\min},y)$ are scaled attribute dominance utility functions of one dimension since the utility function is zero if that single attribute is set to a minimum, and the scaling constants are equal to $U_{xy}(x_{\max},y_{\min})$ and $U_{xy}(x_{\min},y_{\max})$ respectively.

Now we discuss the integral term $I(x,y) = \int_{y_{\min}}^{y} \int_{x_{\min}}^{x} u_{xy}(x,y)dxdy$. We note that, by properties of a definite integral, $I(x_{\min},y) = I(x,y_{\min}) = 0$. In other words, this integral has a value of zero if either $x$ or $y$ is at the minimum value. Furthermore, this integral has a maximum value equal to $k_{xy} = I(x_{\max},y_{\max})$. Let us normalize this integral term and define $C(x,y)$ as

$$C(x,y) \triangleq \frac{\int_{y_{\min}}^{y}\int_{x_{\min}}^{x} u_{xy}(x,y)dxdy}{\int_{y_{\min}}^{y_{\max}}\int_{x_{\min}}^{x_{\max}} u_{xy}(x,y)dxdy} = \frac{I(x,y)}{I(x_{\max},y_{\max})} \qquad (22)$$

From equation (22), we note that $C(x,y)$ has the same mathematical properties as an attribute dominance utility function. In other words we can write

$$U_{xy}(x,y) = k_x C(x,y_{\min}) + k_y C(x_{\min},y) + k_{xy} C(x,y) \qquad (23)$$

where $k_x, k_y$, and $k_{xy}$ are scaling constants and $C(x,y_{\min}), C(x_{\min},y),$ and $C(x,y)$ are normalized attribute dominance utility functions.

Similarly, for three dimensions we can define

$$u_{xyz}(x,y,z) \triangleq \frac{\partial^3 U(x,y,z)}{\partial x \partial y \partial z}, \tag{24}$$

and therefore,

$$\begin{aligned}U_{xyz}(x,y,z) &= U_{xyz}(x,y_{\min},z_{\min}) + U_{xyz}(x_{\min},y,z_{\min}) + U_{xyz}(x_{\min},y_{\min},z) \\ &- U_{xyz}(x_{\min},y,z) - U_{xyz}(x,y_{\min},z) - U_{xyz}(x,y,z_{\min}) \\ &+ \int_{z_{\min}}^{z}\int_{y_{\min}}^{y}\int_{x_{\min}}^{x} u_{xyz}(x,y,z)dxdydz\end{aligned} \tag{25}$$

Once again we note that the terms $U_{xyz}(x,y_{\min},z_{\min})$, $U_{xyz}(x_{\min},y,z_{\min})$, and $U_{xyz}(x_{\min},y_{\min},z)$ are scaled attribute dominance utility functions of one dimension, $C(x,y,z) = \int_{z_{\min}}^{z}\int_{y_{\min}}^{y}\int_{x_{\min}}^{x} u_{xyz}(x,y,z)dxdydz$ is a scaled attribute dominance utility function of three dimensions, and that $U_{xyz}(x_{\min},y,z)$, $U_{xyz}(x,y_{\min},z)$, and $U_{xyz}(x,y,z_{\min})$ can be reduced to one dimensional and two dimensional attribute dominance utility functions using the arguments for the two dimensional case presented above.

In other words, we can write equation (25) as

$$\begin{aligned}U_{xyz}(x,y,z) &= k_x C(x,y_{\min},z_{\min}) + k_y C(x_{\min},y,z_{\min}) + k_z C(x_{\min},y_{\min},z) \\ &- k_{yz} C(x_{\min},y,z) - k_{xz} C(x,y_{\min},z) - k_{xy} C(x,y,z_{\min}) \\ &+ k_{xyz} C(x,y,z)\end{aligned} \tag{26}$$

Equation (26) expresses any general three-dimensional multiattribute utility function in terms of a linear combination of attribute dominance utility functions. By induction, we can show that the same formulation applies to higher dimensions.

We note that a special case of decomposing utility functions into attribute dominance utility functions arises when the attributes have mutual utility independence and leads to the multilinear utility function. In order to see this, we recall the following theorem from about mutual utility independence [8].

If two attributes, $x$ and $y$, have mutual utility independence, their multiattribute utility function has the multilinear form

$$U_{xy}(x,y) = k_y U_y(y) + k_x U_x(x) + k_{xy} U_x(x) U_y(y) \tag{27}$$

where $k_y$, $k_x$, and $k_{xy} = 1 - k_x - k_y$ are constants, and $U_y(y)$ and $U_x(x)$ are the utility functions of attributes $y$ and $x$ respectively. If we substitute for $C(x,y) = U(x)U(y)$ into (23) we note that (27), is indeed a special case.

# MAXIMUM ENTROPY ATTRIBUTE DOMINANCE UTILITY FUNCTIONS

By analogy with joint probability density functions, we can now we extend the maximum entropy utility analysis using the joint utility density function that we have developed. For example, if the decision maker can provide only the marginal utility function for each attribute present, then the maximum entropy formulation maximizes the entropy of the joint utility density function subject to the given preference information constraints. If for example, we have two attributes $x$ and $y$ that are defined on the unit square, and we know the marginal utility functions, the formulation for the problem is

$$\max \; -\int_1^1\int_0^1 u(x,y)\ln(u(x,y))dxdy$$

st

$$\int_0^1\int_0^x u(x,y)dxdy = U_x(x), \tag{28}$$

$$\int_0^y\int_0^1 u(x,y)dxdy = U_y(y),$$

$$\int_0^1\int_0^1 u(x,y)dxdy = 1.$$

The solution to this problem provides a joint utility density function that is equal to the product of the marginal utility density functions.

$$u(x,y) = u_x(x)u_y(y), \quad 0 \leq x \leq 1, 0 \leq y \leq 1. \tag{29}$$

The joint utility density function integrates, using equation (20), to

$$U(x,y) = U_x(x)U_y(y), \quad 0 \leq x \leq 1, 0 \leq y \leq 1. \tag{30}$$

The product form of the utility density functions presented in equation (29) is the condition of utility independence that is analogous to probability independence and applies to any general class of multiattribute utility functions that have mutual utility independence. We can show this by substituting equation (27) into equation (18) to get

$$u(x,y) \triangleq \frac{\partial^2 U(x,y)}{\partial x \partial y} = k_{xy}u_x(x)u_y(y) \tag{31}$$

If the marginal utility functions are known, the maximum entropy attribute dominance utility function assumes utility independence between the attributes. If more information is available about the utility function, such as moments, risk aversion coefficients, or utility dependence parameters, it can also be incorporated into the maximum entropy formulation and an attribute dominance utility function can be obtained by integration. We recall that the construction of any general multiattribute utility function involves the construction of several attribute dominance utility functions of different dimensions.

# RELATIVE ENTROPY

The KL-distance (often known as the relative entropy or cross-entropy) is a measure of divergence between two probability distributions. In a similar manner, the KL-distance is a measure of the divergence between two utility density functions.

The expression for the KL-distance between a utility density functions, $u(x)$ and another utility density function $q(x)$ is given as

$$KL - distance = \iint u(x) \ln(\frac{u(x)}{q(x)}) \qquad (32)$$

The maximum entropy utility principle can now be generalized by analogy to a minimum cross-entropy utility principle by minimizing the KL-distance relative to a known target utility density function. This helps incorporate information about the shape of the utility function or its relation to a family of utility functions into the maximum entropy formulation.

# UTILITY DEPENDENCE: AN INFORMATION THEORETIC POINT OF VIEW

In the previous sections we discussed the interpretation of the entropy and relative entropy as they apply to utility functions. In this section we discuss the interpretation of applying several measures of dependence that were developed in information theory into measurements of utility dependence. We provide some insights on their dual interpretations and an intuitive meaning for utility dependence that results from the analogy with joint probability functions. We start with the notion of conditional entropy.

## Conditional Entropy

Having discussed a joint utility density function, and a conditional utility function, we can now define the conditional entropy of attribute Y given a guaranteed amount of attribute X as

$$h(Y | X) = -\iint u(x, y) \ln(u(y | x)), \qquad (33)$$

where $u(y|x) = \frac{\partial}{\partial y} U(y|x)$ is the conditional utility density function of *y* for a given value of *x*.

It is known from information theory that the conditional entropy is less than or equal to the marginal entropy, with equality if and only if the two variables are independent. I.e.

$$h(Y|X) \leq h(Y). \tag{34}$$

The interpretation for this result when applied to utility theory is that our utility functions for attributes have less entropy (become closer to aspiration utility functions) when conditioned on another attribute. However, when the two attributes are utility independent, the conditional entropy remains the same as the marginal entropy, since the utility functions for each attribute are not updated.

## Mutual Preference

The notion of mutual information translates using the utility-probability analogy into mutual preference. In probability land, the mutual information between two variables is the KL-distance between their joint distribution and the product of their marginal distributions. The interpretation for the mutual information is the amount of information induced about a random variable, X, when we know the outcome of another random variable, Y.

Let us now consider the two extreme cases of probabilistic dependence. The first case is when the two variables are independent. In this case we have a marginal probability distribution for each variable. Knowing the outcome of one variable, X, tells us nothing about the other variable Y, and so this knowledge does not update its probability distribution. Therefore the entropy of Y remains unchanged whether or not we know X. The second case is when the two variables are deterministically related through a functional expression. For example, Profit = 0.15*Revenue. In this case, knowledge of the Revenue tells us the exact value of profit. If we know revenue, the distribution of profit is updated to a step distribution function (or an impulse probability density function) at that particular value of profit. The entropy of the updated distribution of X is reduced to negative infinity (it is now deterministic) once we know the outcome of Y. The mutual information between the two variables in this case is infinite as knowledge of one variable provides infinite reduction in the entropy of the other.

Now we consider the analogous interpretation of mutual information when applied to utility, which we call mutual preference. If we have a prospect with two attributes, we have a utility function for each attribute. Now we consider two extreme cases. The first is when the attributes have utility independence. If we are now guaranteed a fixed amount of one attribute, it does not change our utility function over the other (by definition of utility independence), and therefore this guarantee does not change its entropy. The mutual preference between the attributes is therefore equal to zero. On the other hand, the attributes may be deterministically related. Once again, we consider the example Profit = 0.15*Revenue. In this case, if we are guaranteed a fixed amount of Revenue, say $R_0$, then we automatically set a reference point for our satisfaction with profit, $0.15*R_0$. Any value of profit that is below $0.15*R_0$ will be unsatisfactory for us while any value of profit that is above $0.15*R_0$ will be satisfactory. In other words, we now have a step (aspiration) utility function (or an impulse utility density function) for profit. A guarantee of the Revenue attribute has sharpened our utility function for profit and it is now a target or aspiration level. The mutual preference between the two attributes is thus infinite since guarantee of one attribute reduces the entropy in the utility function of the other to its minimum value of negative infinity.

# AXIOMATIC JUSTIFICATION OF THE MAXIMUM ENTROPY UTILITY PRINCIPLE

We have discussed the maximum entropy utility principle and the application of several measures of information theory to utility theory. In this section we further draw on this analogy to provide an axiomatic derivation for the maximum entropy utility principle. Shore and Johnson [9] showed that Jaynes' maximum entropy principle for probability inference satisfies a set of reasonable axioms that stem from one fundamental principle: if a problem can be solved in more than one way, the results should be consistent. They also showed that maximizing any function but the entropy will lead to inconsistencies in these axioms unless that function and the entropy have identical maxima (any monotonic function of the entropy will work). The analogy between probability and utility density functions that we have developed allows the application of their analysis directly to the maximum entropy utility principle. Now we will mention each of the axioms and discuss how they relate to utility assignment.

## Axiom 1: Uniqueness

The utility values that we assign by maximum entropy utility should be unique. The justification for this is that if we do not have unique utility values, we are vulnerable to being a "money pump" (as discussed earlier). Another justification for the uniqueness of the maximum entropy utility assignment is that if we solve the utility assignment problem twice in exactly the same way, we expect the same utility values to be assigned in both times.

## Axiom 2: Invariance

We should expect the same utility values from the maximum entropy utility principle when we solve the problem in two different coordinate systems. For example, if a utility density function is assigned in Cartesian co-ordinates and integrate to get a multiattribute utility function, we should get the same results if we solve the problem in polar coordinates and integrate over the domain of the polar coordinates to get a utility function.

## Axiom 3: System Independence

Consider two independent utility functions or two attributes that have utility independence. If we receive information about the two attributes, then we would like the joint utility density function that is assigned to be equal to the product of the individual utility density functions (implying utility independence as discussed above).

## Axiom 4: Subset Independence

This axiom concerns itself with situations where we receive preference information about the conditional utility functions in attribute dominance utility. One way of accounting for this new information is to update the conditional utility functions $U(B|A)$ individually. Another way is to marginalize the attribute dominance utility function to get $U(B)$ and then apply the new preference information to $U(B)$. The results should be consistent in both cases.

As mentioned above, if these axioms are required by a utility assignment mechanism, then maximizing any function but the entropy will lead to inconsistencies in these axioms unless that function and the entropy have identical maxima.

In addition to satisfying the previous axioms, the maximum entropy utility principle satisfies two essential desiderata. The first desideratum, as mentioned earlier, is that for different people with the same order of preferences for the prospects and where we have the same preference information we should assign the same utility values. The second desideratum is -utility and probability independence- that results from the foundations of normative utility theory. The utility value of a prospect should not depend on the probability of getting that prospect due to the normative separation of beliefs from preferences [10].

## CONCLUSIONS

In this paper we have provided an analogy between probability and utility and built on the notion of a utility density function to a joint utility density function for multiple attributes. We applied concepts of information theory to utility theory and gained insights on the notions of utility dependence and aspiration utility functions. The analogy between probability and utility provides several directions for future research. For example, analogous interpretations for Cox's axioms provide the basis for an axiomatic derivation of utility inference. The algebra of probable inference translates to the algebra of preferences for logical operations on the attributes. The idea of Influence diagrams as graphical representations of joint probability distributions [11] extends to the notion of utility diagrams, which are graphical representations of multiattribute utility functions and express the dependence relations between the attributes. The method of copulas for constructing joint cumulative distribution functions translates into a method for constructing attribute dominance utility where the dependence parameters can be determined by the trade-off coefficients between the attributes. Assessing utility moments and utility cross-correlation coefficients can also be used to construct a multiattribute utility function. Interchanging utility functions with probability functions provides a wealth of new results and insights. For example, the notion of utility dominance is the analog of stochastic dominance [12].

# ACKNOWLEDGEMENTS

I would like to thank Ronald Howard and Myron Tribus for introducing me to the Maximum Entropy community and to the work of Edwin Jaynes and Richard Cox.